# Out-of-focus Blur: Image De-blurring

Yuzhen Lu

Department of Biosystems and Agricultural Engineering

Michigan State University, East Lansing, MI 48824

524 S. Shaw Lane, 105A A.W. Farrall Hall

(May 6, 2016)

**Abstract** Image de-blurring is important in many cases of imaging a real scene or object by a camera. This project focuses on de-blurring an image distorted by an out-of-focus blur through a simulation study. A pseudo-inverse filter is first explored but it fails because of severe noise amplification. Then Tikhonov regularization methods are employed, which produce greatly improved results compared to the pseudo-inverse filter. In Tikhonov regularization, the choice of the regularization parameter $\mu$ plays a critical rule in obtaining a high-quality image, and the regularized solutions possess a semi-convergence property. The best result, with the relative restoration error of 8.49%, is achieved when the prescribed discrepancy principle is used to decide an optimal $\mu$ value. Furthermore, an iterative method, Conjugated Gradient, is employed for image de-blurring, which is fast in computation and leads to an even better result with the relative restoration error of 8.22%. The number of iteration in CG acts as a regularization parameter, and the iterates have a semi-convergence property as well.

## 1. Introduction

Images taken by a camera, in general, suffer from certain level of degraded representation of the original object during the process of image formation. Blur is a common, unwanted artifacts associated with image formation. Blurring is a deterministic process, which can occur due to numerous reasons, such as atmospheric distortions motion, optical aberrations, motion, etc. [1]. When a real scene is imaged by a camera, some of the points are in focus while others not, thus causing out-of-focus blurring. Out-of-focus blurring is space-invariant in cases where the surface of a flat object is parallel to the image plane [2]. In addition to blurring, image degradation is also caused by noise during image recording.

Mathematically, image degradation due to blurring and noise can be modelled by a linear system as follows:

$$g(x, y) = Af(x, y) + n(x, y) \tag{1}$$

where $(x, y)$ represents spatial coordinates, $g(x, y)$ is the acquired image with blur and noise, $A$ is the blurring matrix defined by a certain point spread function (PSF), $f(x, y)$ is a true image and $n(x, y)$ is noise. In terms of PSF, Eq. (1) can be rewritten as follows:

$$g(x, y) = K(x, y) \otimes f(x, y) + n(x, y) \tag{2}$$

where $K(x, y)$ is the PSF of the imaging system that causes image blurring, and $\otimes$ denotes two-dimensional (2-D) convolution. Each kind of blurring has a corresponding PSF. The PSF of the space-invariant or shift-invariant out-of-focus blurring, e.g., can be described by [3]:

$$K(x, y) = \begin{cases} \dfrac{1}{\pi r^2} & \text{if } (x-k)^2 + (y-l)^2 \leq r^2, \\ 0 & \text{elsewhere,} \end{cases} \tag{3}$$

Where $(k, l)$ is the center of the PSF, and $r$ is the radius of the blur.

Image de-blurring is an inverse problem that is to reconstruct the true, sharp image by solving Eq. (1) or (2) for $f(x, y)$. It is noted that there is no hope to restore the true image exactly due to the loss of some information during image formation and various unavoidable measurement errors. Despite this, one can recover as much information as possible from a given image by developing effective and reliable algorithms aided with knowledge of the blurring process. The following discussion of this report is concentrated on image de-blurring due to shift-invariant out-of-focus blur. Section 2 is to give an overview of image de-blurring techniques; Section 3 is to present a simulation study to investigate the performance of different image de-blurring methods, followed by conclusions in Section 4.

## 2. Image de-blurring

Numerous methods exist for image de-blurring, which basically can be divided into non-blind and blind deconvolution [4], depending on if the blurring kernel is known or not. This report is only focused on non-blind image de-blurring, which assumes that the blur kernel is *in prior* known, to be specific, including pseudo-inverse filtering, regularization methods and conjugate gradient.

### 2.1. Pseudo-inverse filtering

A pseudo-inverse filter is a linear filter whose PSF is the inverse of the blurring kernel. Given a linear blurring model as described in Eq. (2), applying Fourier transform yields:

$$g(u, v) = \hat{K}(u, v) \hat{f}(u, v) + n(u, v) \tag{4}$$

where $(u, v)$ are the coordinates in the Fourier space; $\hat{g}(u,v)$, $\hat{K}(u,v)$, $\hat{f}(u,v)$ and $\hat{n}(u,v)$ are Fourier transformations of $g(u,v)$, $K(u,v)$, $f(u,v)$ and $n(u,v)$, respectively. The pseudo-inverse filter is to divide this equation by $\hat{K}(u,v)$ and get $\hat{F}(u,v)$ as an estimate of $\hat{f}(u,v)$ of the true source image:

$$\hat{F}(u,v) = \hat{f}(u,v) + \frac{\hat{n}(u,v)}{\hat{K}(u,v)} \tag{5}$$

Then, the true image is derived by taking the inverse Fourier transform of $\hat{F}(u,v)$. The pseudo-inverse filter gives the simplest solution to the de-blurring problem. However, it in practice almost never works unless the acquired image is noise-free. The magnitude response of the blur has some very low values. When performing division pointwise, one is also dividing the noise term by these same low values, thus leading to a huge amplification of noise that is enough to completely swamp the image itself. This will be demonstrated in the simulation study.

## 2.2. Regularization Methods

Most of de-blurring methods are using certain type of regularization to get rid of the issue of noise amplification in the final solution. The basic ideal of regularization lies in enforcing certain regularity conditions on the solution, and the degree of regularization depends on a regularization parameter. Two candidate regularization methods are available, including truncated singular value decomposition (TSVD) and Tikhonov regularization [3, 5]. The following discussion will be only restricted to the latter.

In the Fourier space, the Tikhonov regularization can be formulated as follows [2]:

$$\hat{F}_\mu(u,v) = \frac{\hat{K}^*(u,v)}{\left|\hat{K}(u,v)\right|^2 + \mu} \hat{g}(u,v) \tag{6}$$

where $\hat{K}^*(u,v)$ is the complex conjugate of $\hat{K}(u,v)$ and $\mu$ is the regularization parameter. The choice of $\mu$ yields the solution to the minimization problem as follows [2]:

$$\min_f \|Af(x,y) - g(x,y)\|^2 + \mu \|f(x,y)\|^2 \tag{7}$$

The Tikhonov regularization is equivalent to a filtering of the pseudo-inverse solution, which is a spectral filtering method, in a broad sense.

The choice of the regularization parameter $\mu$ is critical for attaining a desirable result. The regularization parameter can be decided based on some criteria including the prescribed energy, the prescribed discrepancy, the Miller's method, generalized cross-validation (GCV), the L-curve, etc. [2]. In this report, the choice of regularization parameter µ is based on the first four criteria, in which µ satisfies the following four equations respectively:

$$\|f_\mu(x,y)\| = E \tag{8}$$

$$\|Af_\mu(x,y) - g(x,y)\| = \varepsilon \tag{9}$$

$$\mu = (\varepsilon/E)^2 \tag{10}$$

$$V(\mu) = \frac{\|Af_\mu(x,y) - g(x,y)\|^2}{\left\{Tr\left[I - AA^T\left(AA^T + \mu I\right)^{-1}\right]\right\}^2} \tag{11}$$

where $E$ and $\varepsilon$ are the prescribed energy and discrepancy respectively; $Tr$ is the trace operator. Note that $E$ and $\varepsilon$ are computed by taking the norms of the true image and the additive random noise respectively. For Eq. (11), in particular, the optimal regularization parameter is the minimizer of the function $V(\mu)$ that is unique. To use GCV in the Fourier space to estimate the regularization parameter, we further simply Eq. (11) as follows [3]:

$$V(\mu) = \frac{\sum_{i=1}^{N}\left(\frac{\hat{g}(u,v)}{s_i^2 + \mu}\right)^2}{\sum_{i=1}^{N}\left(\frac{1}{s_i^2 + \mu}\right)^2} \tag{12}$$

where $s$ is the magnitude values of $\hat{K}(u,v)$, corresponding to the singular values of the blurring matrix $A$; and $N$ is total number of image pixels.

## 2.3. Conjugate Gradient

Iterative methods have also been utilized to solve ill-posed problems including image de-blurring. Compared to direct regularization, iterative methods are advantageous due to efficient computation, especially for large matrices, and they can incorporate a variety of regularization techniques and can more easily incorporate additional constraints [6-8]. In this report, special attention is paid to the Conjugate Gradient (CG) method that is relatively fast in image restoration.

In the context of Fourier transform, the CG method can be implemented following the algorithms presented as below [2]:

$$f_0 = 0 \,;\, r_0 = p_0 = K^*g \tag{13}$$

$$\alpha_k = \frac{\sum_{k=1}^{N} |\hat{r}_k|^2}{\sum_{k=1}^{N} |\hat{K}^*|^2 |\hat{p}_k|^2} \qquad (14)$$

$$r_{k+1} = r_k - \alpha_k |\hat{K}|^2 p_k \qquad (15)$$

$$\beta_k = \frac{\sum_{k=1}^{N} |r_{k+1}|^2}{\sum_{k=1}^{N} |\hat{r}_k|^2} \qquad (16)$$

$$p_{k+1} = r_{k+1} + \beta_k p_k \qquad (17)$$

$$f_{k+1} = f_k + \alpha_k \hat{p}_k \qquad (18)$$

For brevity, the Fourier space coordinates ($u$, $v$) are omitted in Eqs. (13-18). Since the iterates $\hat{f}_k$ have the semi-convergence property, i.e., the $\hat{f}_k$ first approach to the desired solution and then go away, and hence the prescribed discrepancy principle can be used as a stopping criterion to regulate the iteration process.

## 3. Simulation Study

### 3.1. Image blurring

An eight-bit gray-scale image of the dimension 512×512 is experimented as a true image for image blurring and de-blurring. An out-of-focus PSF of the same size is created according to Eq. (3) with the blur radius set to 15. The blurred image is generated by convolving the PSF with the true image, and then is contaminated by Gaussian random noise with the signal-to-noise ratio of 40 dB. The image processing throughout the simulation study is implemented in Maltab 2014b. Fig. 1 shows the true image, PSF and the blurred results.

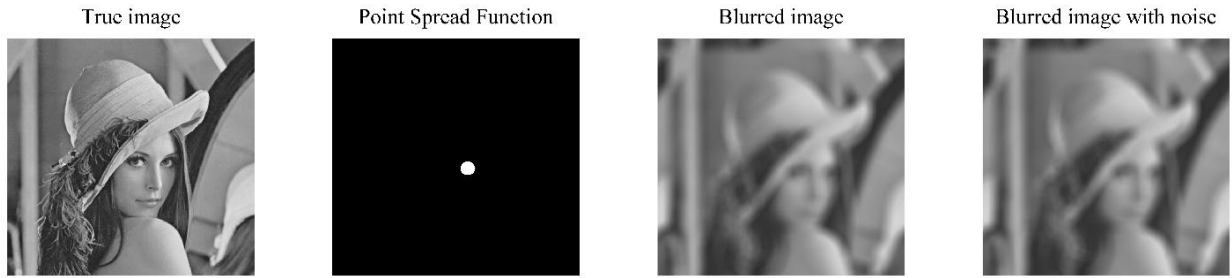

Fig. 1 Example of image blurring

### 3.2. Image de-blurring

#### 3.2.1. Pseudo-inverse filtering

The pseudo-inverse filtering is first employed to restore the blurred image. Fig. 2 (left) shows the de-blurred image. Obviously, the method fails, and it achieves nothing but noise. The reason is rather clear when we look at the Fourier magnitude map of the blurring PSF (Fig. 2, right). The magnitude map is sort of a 2-D impulse function whose magnitude responses away from the center peak are close to zero. When we divide the blurred image by the PSF in the Fourier space, we are also dividing the noise term present in the blurred image by the same very low values, thus greatly amplifying the contribution to the restoration from the noise component. Hence, the noise dominates in the resultant image.

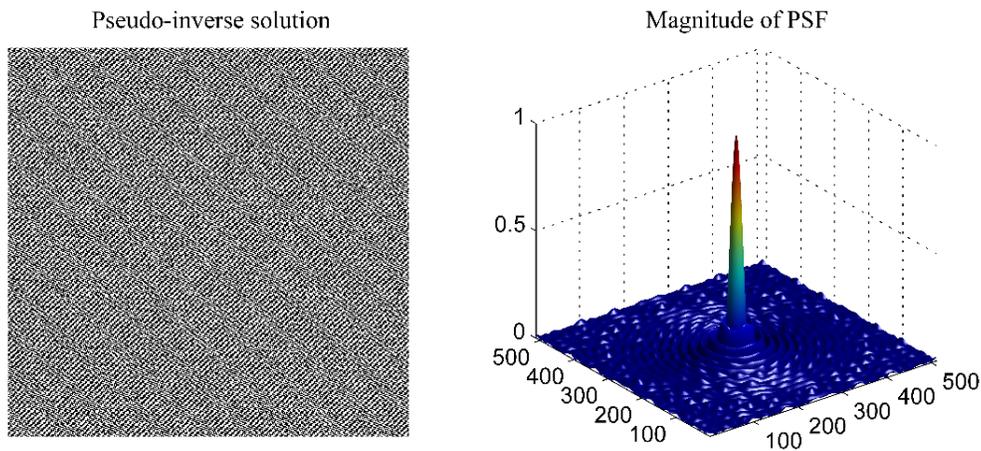

Fig. 2 The de-blurred image by pseudo-inverse filtering (left) and FFT magnitude map of PSF (right)

### 3.2.2. Tikhonov Regularization

Now regularization methods are employed to restore the original image. First, the effect of the regularization parameter μ on the de-blurring performance is investigated by varying it from 1e-7 to 10. Fig. 3 shows the restored images. It can be seen that as the μ increases, the recovered image becomes clear and sharp, with reduced noise and well-resolved image features; but when μ goes up further, the image becomes smooth and obscure until no features cannot be recognized. This phenomena demonstrate a semi-convergence property of the regularized solution with varying regularization parameters in the cases of noisy images [2]. Underestimating the parameter tends would cause severe noise contamination while overestimating leads to much loss of image details. So, there should exist an optimal value of the regularization parameter which can achieve a trade-off between noise level and image details, and give the lowest restoration error.

Furthermore, the behavior of the relative restoration error versus the regularization parameter is examined. As shown in Fig. 4, the relative restoration error decreases as the μ increases until μ is equal to 0.001 after which the error rises up. Hence the best value of the regularization parameter is around 0.001. This findings accord with visual inspection of de-blurred results.

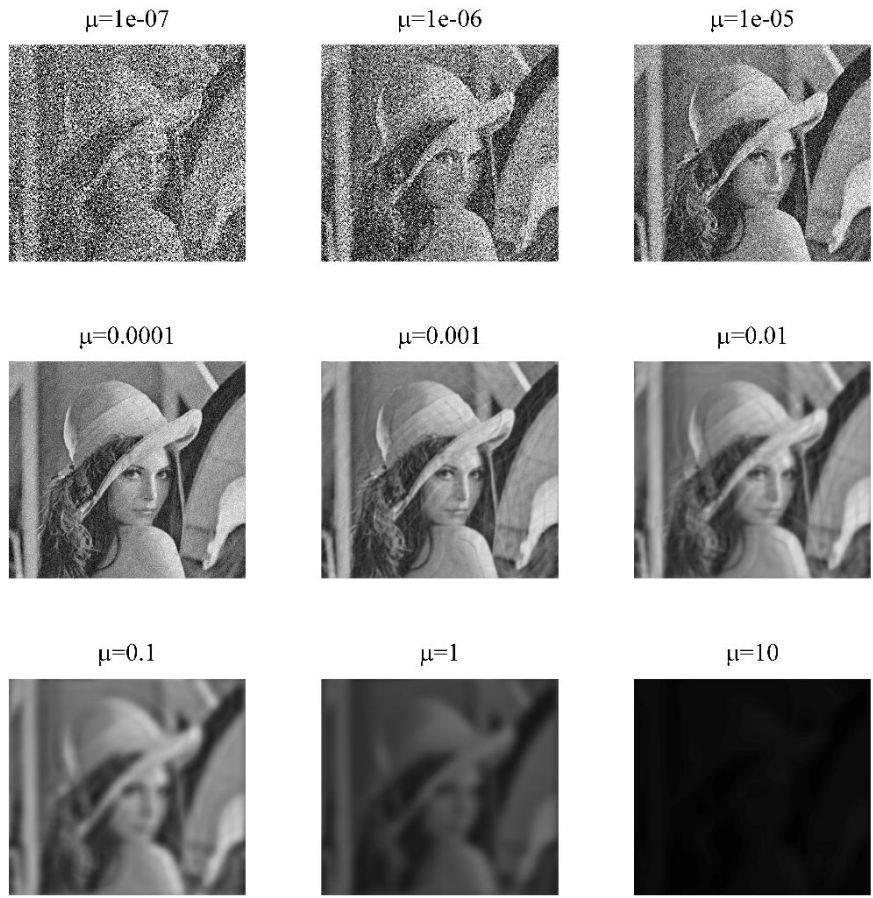

Fig. 3. Effect of the regularization parameter

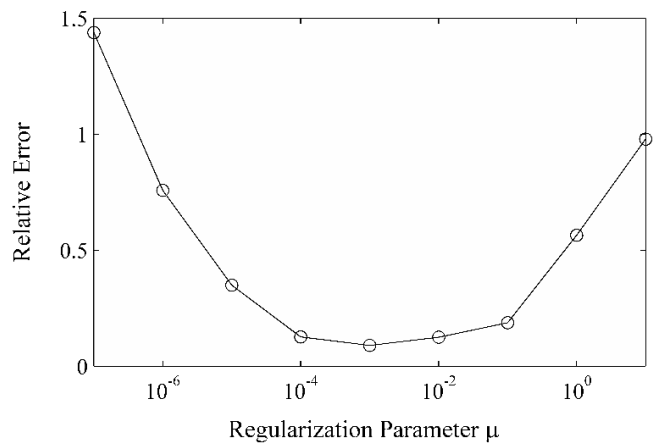

Fig. 4. Relative restoration error versus the regularization parameter

Next, the regularization parameter is chosen based on the prescribed energy, the prescribed discrepancy, the Miller's method and generalized cross-validation (GCV) respectively. Fig. 5 shows the de-blurred results. All the images are far better in quality than that obtained by the pseudo-inverse filtering; apparently, the prescribed discrepancy principle yields the best image, followed by the GCV and the prescribed energy, and the Miller's method is the worst because of obvious noise residual.

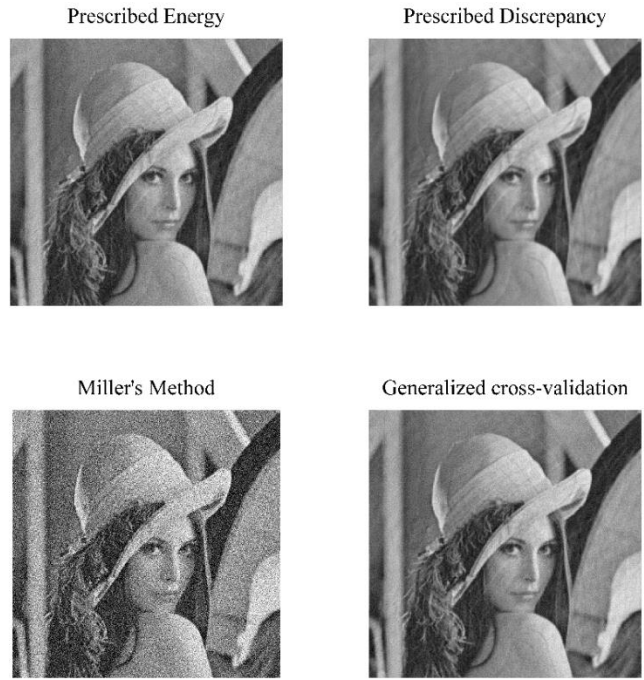

Fig. 5 De-blurred methods by regularization methods

Furthermore, the relative restoration error is calculated to compare the four regularization ways. As summarized in Table 1, the prescribed discrepancy method is the best due to the lowest error, followed by GCV which is slightly better than the prescribed energy, and the Miller's method results in the largest error. These results are consistent with visual inspection above.

Table 1 The μ values and the relative restoration errors by different regularization methods

| Regularization methods | Prescribed energy | Prescribed discrepancy | Miller's method | Generalized cross-validation |
| --- | --- | --- | --- | --- |
| Regularization parameter | 1.54e-4 | 5.00e-4 | 8.21e-6 | 1.73e-4 |
| Relative restoration error | 10.54% | 8.49% | 37.65% | 10.11% |

The prescribed discrepancy method obtains a regularization parameter of 5.00e-4, which is the closest to 0.001 revealed by Fig. 4, and the GCV and the prescribed energy also gives reasonable estimate of the regularization parameter and thus respectable de-blurring results. The Miller's method, although it has an explicit and straightforward formula for the regularization parameter, usually tends to underestimate the value of the regularization parameter and hence suffers from an elevated high level of noise contamination [2].

### 3.2.3. Conjugated gradient (CG)

Different from the methods above, the CG is an iterative method, in which the number of iterations (k) plays a critical rule in finding a sensible approximate solution [2]. Fig. 6 shows the effect of k on the de-blurring performance. After several steps, the method is able to yield a decent result. As the k increases, the de-blurring performance is improved; but after certain iterations, the restored image deteriorates due to aggravated noise corruption.

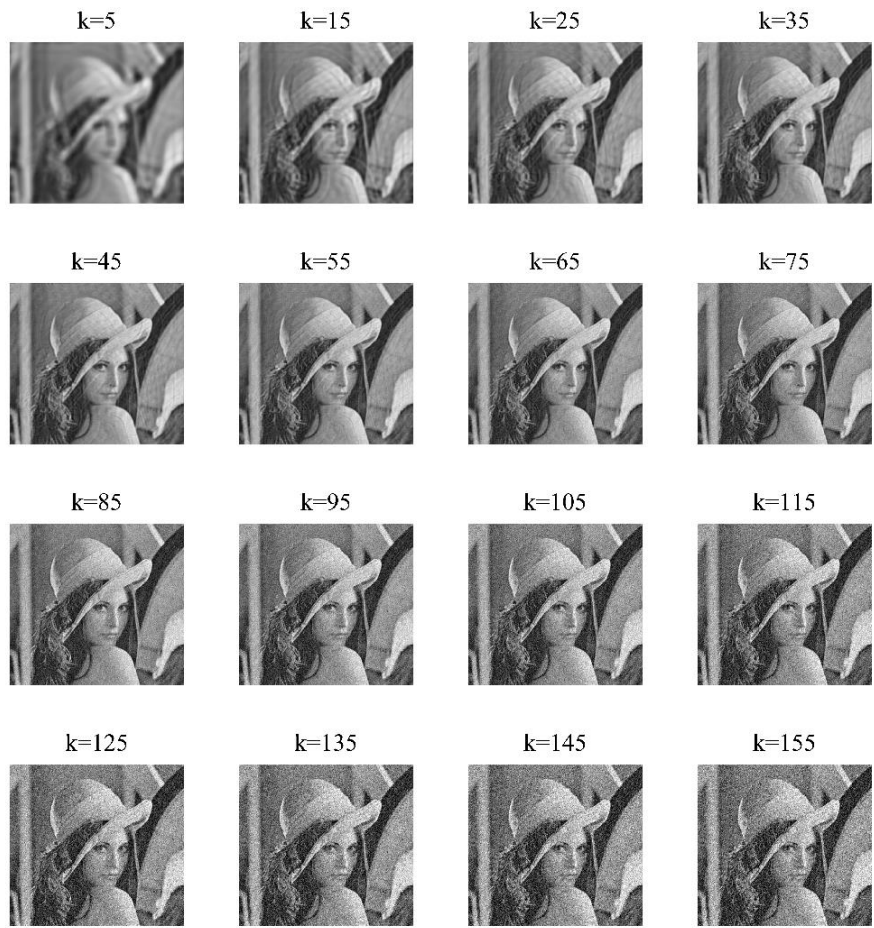

Fig. 6 The effect of the number of iterations on image de-blurring

Fig. 7 presents the behavior of the relative restoration error as a function of the number of iterations k. Obviously, the iterates of the CG method possess a semi-convergence property, as the solutions of Tikhonov regularization methods do. The number of iterations acts as a regularization parameter, and the restoration error has a minimum at a certain value of k, around 50 based on the plot of Fig. 7. Here we assume that *k* is equal to 50 that may not be globally optimal, resulting in a relative restoration error is 8.22%. The error is the lowest compared to the results obtained above by other methods. It is noted that the optimal *k* value can also be decided straightforwardly by means of the discrepancy rule (not presented).

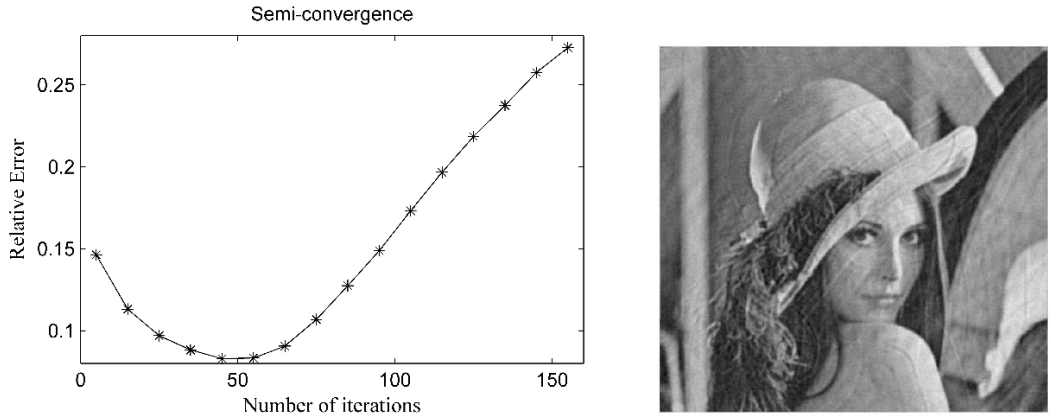

Fig. 7 Left: The relative restoration error of CG versus the number of iterations; right: CG de-blurring with 50 iterations and the corresponding relative restoration error of 8.22%

## 4. Conclusions

This project reports on de-blurring the image that is degraded by the out-of-focus blur and noise, by means of different methods, including pseudo-inverse filtering, Tikhonov regularization and conjugated gradient (CG). The pseudo-inverse filtering gives nothing but noise, because the noise component in the blurred image has been seriously amplified, thus completely swamping the image itself. Regularization methods greatly improve the de-blurred results. The regularization parameter $\mu$ plays a crucial rule in obtaining a desirable result, and the regularized solutions possess a semi-convergence property. Among the four criteria used to choose an optimal $\mu$, the prescribed discrepancy principle yields the lowest relative restoration error of 8.49%, followed by generalized cross validation (GCV) and the prescribed energy; and the Miller's method is the worst due to the underestimated $\mu$ value and thus aggravated noise corruption. The CG is an iterative method, which is fast in computation and also achieves the best result in terms of the restoration error of 8.22%. In implementing CG, the number of iteration has the regularization effect, and its solutions have a semi-convergence property as well.